\pdfoutput=1
\documentclass[11pt]{article}

\usepackage[preprint]{acl}

\usepackage{times}
\usepackage{latexsym}
\usepackage{threeparttable}

\usepackage[T1]{fontenc}
\usepackage[utf8]{inputenc}

\usepackage{microtype}

\usepackage{inconsolata}

\usepackage{booktabs}
\usepackage{multirow}
\usepackage{amsmath}
\usepackage{amssymb}
\usepackage{graphicx}
\usepackage{subfigure}

\usepackage{float}

\usepackage{subcaption}
\usepackage{url}

\newenvironment{itemize*}%
 {\leftmargini=10pt\begin{itemize}%
  \setlength{\itemsep}{0pt}%
  \setlength{\parskip}{0pt}%
  }%
 {\end{itemize}}
\newenvironment{enumerate*}%
 {\begin{enumerate}%
  \setlength{\itemsep}{0pt}%
  \setlength{\parskip}{0pt}}%
 {\end{enumerate}}

\title{Chain of Thought Explanation for Dialogue State Tracking}

\author{Lin Xu$^1$, Ningxin Peng$^2$, Daquan Zhou$^2$, See-Kiong Ng$^1$, Jinlan Fu$^1$\\
$^1$National University of Singapore, $^2$ByteDance \\
\texttt{\{cathyxl2016,zhoudaquan21,jinlanjonna\}@gmail.com}\\
\texttt{seekiong@nus.edu.sg}
\\}
\begin{document}
\maketitle
\begin{abstract}
Dialogue state tracking (DST) aims to record user queries and goals during a conversational interaction achieved by maintaining a predefined set of slots and their corresponding values. 
% However, an intricate challenge arises in the dynamic nature of conversations, where the values assigned to these slots are subject to change over the course of the dialogue, potentially influenced by distant turns in the conversation history. Failure to trace this evolving state can result in erroneous capture of the dialogue's context.   
Current approaches decide slot values opaquely, while humans usually adopt a more deliberate approach by collecting information from relevant dialogue turns and then reasoning the appropriate values.
In this work, we focus on the steps needed to figure out slot values by proposing a model named \textbf{C}hain-\textbf{o}f-\textbf{T}hought-\textbf{E}xplanation (CoTE) for the DST task. CoTE, which is built on the generative DST framework, is designed to create detailed explanations step by step after determining the slot values. This process leads to more accurate and reliable slot values.
% Based on the generative DST framework, CoTE learns to generate multi-step explanations subsequent to the slot values, which in turn ameliorates more accuracy and reliable slot values. 
Moreover, to improve the reasoning ability of the CoTE, we further construct more fluent and high-quality explanations with automatic paraphrasing, leading the method CoTE-refined. 
% We thoroughly investigate the effectiveness of our CoTE in both full-dataset and low-shot scenarios. 
Experimental results on three widely recognized DST benchmarks-MultiWOZ 2.2, WoZ 2.0, and M2M-demonstrate the remarkable effectiveness of the CoTE. 
Furthermore, through a meticulous fine-grained analysis, we observe significant benefits of our CoTE on samples characterized by longer dialogue turns, user responses, and reasoning steps. 
\end{abstract}
\section{Introduction}
Dialogue State Tracking (DST), as a crucial task of a task-oriented dialogue system, aims to track key information that fulfills users' goals. This key information is maintained in the form of slot-value pairs.
Existing methods update slots by considering the entire dialogue history~\cite{lee:dst-prompt,feng2020seqseq,chen2020schema}, some related turns implicitly~\cite{shin-etal-2022-dialogue,guo:multi-granularity}, 
or solely the current turn~\cite{zhu2020efficient}.
% These approaches may lead to redundant or insufficient information. Consequently, the learned models tend to rely on rules and matching~\cite{jia2017adversarial} rather than developing genuine generalization abilities for state tracking.

\begin{figure}[t]
	\centering
        % \vspace{-10pt}
	\subfigure[An example of reasoning required in DST. ]{
	\begin{minipage}[b]{0.48 \textwidth}
		\includegraphics[width=1\textwidth]{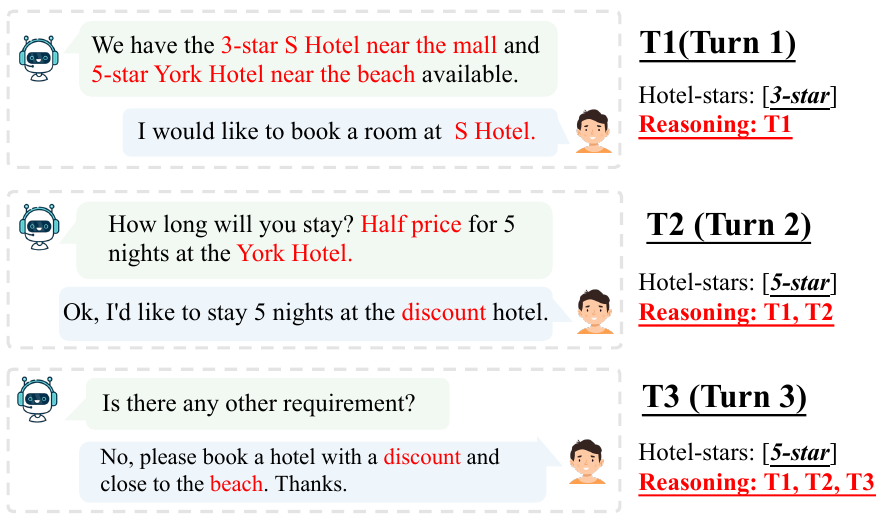}
	\end{minipage}
	} \\
	% \vspace{-10pt}
	\subfigure[Statistics for several popular datasets.]{
	\begin{minipage}[b]{0.48 \textwidth}
		\includegraphics[width=1\textwidth]{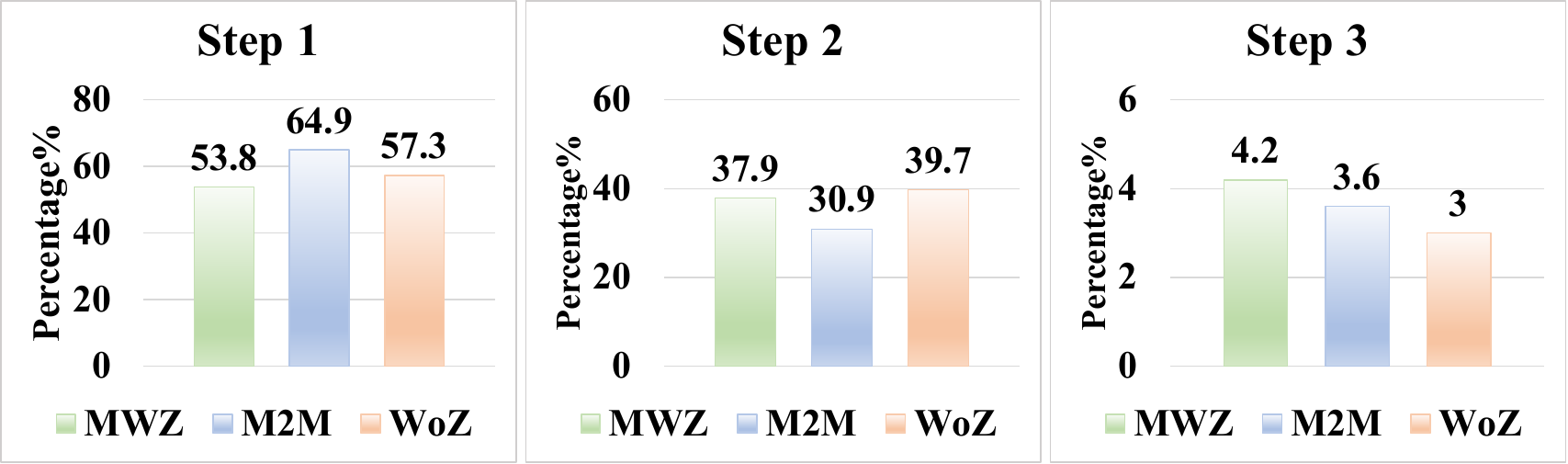}
	\end{minipage}
	} \\
	% \vspace{-10pt}
	% \subfigure[JGA on samples requiring various reasoning steps]{
	% \begin{minipage}[b]{0.44 \textwidth}
	% 	\includegraphics[width=1\textwidth]{fig/full_finegrain.pdf}
	% \end{minipage}
	% }
	\caption{
	(a) A multi-step reasoning example for DST. The slot value of '\textit{hotel-stars}' at turn 3 (T3) depends on the content across T1, T2, and T3.
 (b) Sample ratios from various reasoning steps on MultiWOZ 2.2 (MWZ), M2M, WoZ 2.0 (WoZ) datasets. Nearly 40\% samples require multi-step reasoning ($\text{step}>=2$).
  % that achieve the slot value relies on on several popular datasets, including
 % (c) JGA on samples requiring different reasoning steps
}
    \label{fig:cot-reason}
\end{figure}

\begin{figure}[t]
  \centering
  \includegraphics[width=0.48\textwidth]{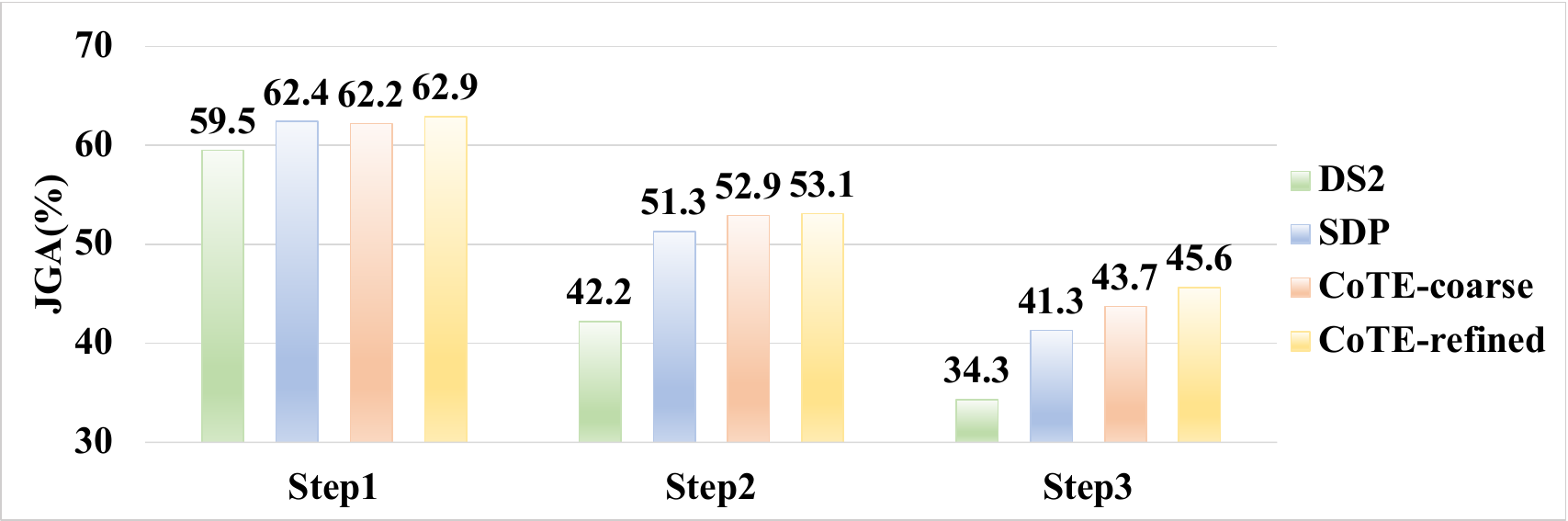}
\caption{Fine-grained analysis on four DST models on  MultiWOZ 2.2 dataset. CoTE-refined outperforms DS2 and SDP on multi-step samples (e.i., step2 and step3) a lot.}
\label{fig:fine-intuition}
\end{figure}

However, tracking slot values in numerous instances necessitates reasoning or information exchange across particular dialogue turns, which may not be adjacent or predetermined in number.
\autoref{fig:cot-reason}-(a) presents an illustrative example. The alteration of slot `\texttt{hotel-stars}' at turn 3 (\texttt{T3}) relies on the logical deduction from the dialogue contents of turn 1 (\texttt{T1}), turn 2 (\texttt{T2}), and turn 3 (\texttt{T3}).
The reasoning steps include:
(1) Step-1: \texttt{T3}; the user decides to stay at a discounted hotel;
(2) Step-2: \texttt{T3,2}; the discounted hotel is York hotel;
(3) Step-3: \texttt{T3,2,1}; York Hotel is s 5-star.
Therefore, the value assigned to `\texttt{hotel-stars}' is \textit{5-star}.

Moreover, as illustrated in \autoref{fig:cot-reason}-(b), a significant proportion (approximately 40\%) of samples from DST's well-established datasets necessitate information spanning multiple dialogue turns to ascertain slot values accurately. This complexity poses a challenge for existing models. \autoref{fig:fine-intuition} provides a fine-grained analysis of several state-of-the-art methods, DS2~\cite{shin-etal-2022-dialogue}, SDP~\cite{lee:dst-prompt}, as well as our CoTE-coarse and CoTE-refined (will introduce in the Q1) models concerning the reasoning steps. The findings reveal that these models excel in performance when confronted with instances involving a single reasoning step (step=1); however, they exhibit diminished performance when dealing with instances requiring multi-step reasoning (step>=2). This observation underscores the limited generalization capacity of these models when applied to complex scenarios in DST.

This work is organized around the following three research questions:
% \textbf{Q1}: Whether accompanying generated explanations assist slot value deduction and show better generalization ability in complex scenarios needing multi-step reasoning? 

\textbf{Q1}: Can introducing reasoning explanations improve performance? 
Generative dialogue state tracking with pre-trained language (PLM) models has been proven to have decent performance~\cite{lee:dst-prompt, shin-etal-2022-dialogue} owing to PLMs' inherent ability to understand the text-based slot names and corresponding slot schema. However, these methods still lack generalization ability according to~\autoref{fig:fine-intuition}. We further claim that adding logical explanations could enhance generative DST models with actual reasoning ability. Chain-of-thought (CoT)~\cite{wei-neurips-2022-chain} as a common practice used in PLMs to handle intricate tasks~\cite{chen:convFinQA,chen:programthoughts} by providing Reasoning~\cite{suzgun:bigbench-cot,chen:table,zhou:least2most} or Explanantions~\cite{stacey:humantouch,marasovic:fewshot,lu:learn2explain}. In this endeavor, we introduce CoT Explanations (CoTE) in DST as a means to mitigate the underperformance observed in samples reliant on multi-step reasoning in relatively small PLMs. 

% \textbf{Q2}: Can the incorporation of fluent and high-quality reasoning explanations further enhance DST tasks?
\textbf{Q2}: Can the high-quality reasoning explanations further improve performance? 
To answer this question, we attempt to leverage GPT3's paraphrasing capabilities to convert segments of dialogue-based reasoning into cohesive third-person narration.

\textbf{Q3}: Where does CoTE improve its performance?
We have devised an interpretable evaluation framework tailored to DST tasks to address this question comprehensively. Our analysis extends across both low-resource and full-dataset settings, allowing us to gain insights into the specific strengths and advantages of CoTE.

The experimental results from both the full dataset and low resource settings demonstrate the following findings:
\begin{itemize*}
    \item The incorporation of the CoT Explanation yields a substantial enhancement in the performance of Dialogue State Tracking.
    \item The utilization of GPT-3 refinement augments the overall performance of DST even further. 
    \item CoTE variants exhibit superior proficiency on intricate samples, for example, with longer turns, utterance length, and reasoning steps, indicating their ability to generalize.
    % that require multi-step reasoning ($\text{reasoning step} >=2$; \autoref{fig:cot-reason}-(c)).
\end{itemize*}
Our primary contributions are as follows:
\begin{itemize*}
    \item We discuss a vital and rarely discussed research problem in the DST task. Statistical analysis shows that about 40\% of DST samples require multi-step reasoning to update the slots, which challenges vanilla models.
    \item We propose CoTE-coarse and its variant CoTE-refined, enhanced by GPT-3, to address the poor performance in samples requiring multi-step reasoning.
    \item To analyze different DST models, we design an interpretable evaluation framework for the DST task and conduct a comprehensive analysis of existing models on the full-dataset and low-resource settings. We provide our code in github link. \url{https://github.com/cathyxl/CoTE-DST} 
\end{itemize*}

\section{Related Work}
\paragraph{Dialogue State Tracking}
Dialogue state tracking aims to track the dialogue state to fulfill user goals in task-oriented dialogue systems. 
Traditional approaches treat DST as a multi-class classification based on a fixed ontology~\cite{henderson-etal-2014-word, zhong-etal-2018-global, chen2020schema}. However, they lack scalability and generalizability due to the inflexible slots/values and the high cost with the increasing number of slots.
Owing to the increasing power of pretrained language models, many recent works~\cite{kim-etal-2020-efficient, feng2020seqseq, guo-etal-2022-beyond, shin-etal-2022-dialogue, cao-2022-d3st} resort to extractive or purely generative techniques to produce slot values. Most of these methods try to suitably choose relevant dialogue context for slot generation~\cite{guo-etal-2022-beyond} or promote efficiency by reusing previously predicted slots~\cite{kim-etal-2020-efficient}. Other works focus on prompt learning to make the most of PLMs' comprehension ability, for example, adapting summarization PLMs in DST~\cite{shin-etal-2022-dialogue}, shuffling slots in prompts to enhance slot understanding~\cite{cao-2022-d3st}. However, none of these methods consider the need for multi-step reasoning in DST.

\paragraph{Pretrained Language Models.} Pretrained Language Models(PLMs) have present incredibly powerful ability when applied in various NLP tasks~\cite{nijkamp2022codegen,scao2022bloom}, especially when model size exceeding hundreds billions~\cite{gpt3-2020-nips, anil2023palm}. To adapt these models to downstream tasks, prompt learning~~\cite{liu2021pre,lester-etal-2021-power} or in-context learning~\cite{gpt3-2020-nips} has been widely applied. The former could be applied in much smaller PLMs with small-scale finetune while the latter is usually used in large language models(LLMs) with LLMs frozen. In this paper, our method is mainly explored with prompt tuning on smaller PLMs, which could also be applied in LLMs, we will leave it as future work. 

\paragraph{Chain-of-Thought Reasoning}
Researchers have found that prompt learning with Chain-of-Thought could elicit the logical and arithmetical ability of the model, thus helping with mathematical problems or question answering~\cite{jason:cot, lu:learn2explain}. More findings further show models could gain the ability to follow a series of logical steps and finally predict the correct answer~\cite{wang-iclr-2023-selfconsistency,chen:programthoughts}. 
% Considering the similar needs in the DST task, we propose a framework that utilizes the idea of Chain-of-Thought to promote the reasoning ability of DST models and boost the performance.

In this work, motivated by Chain-of-Thought Reasoning, we try to construct Chain-of-Thought Reasoning for DST task by extracting multiple system-user utterance pairs from dialogue history that change the slot value.
These utterance pairs are discontinuous yet follow a temporal sequence, thereby exhibiting certain logical relations. We concatenate them, forming Chain-of-Thought Reasoning that comprises multiple reasoning steps. 
We follow previous works~\cite{stacey:humantouch,marasovic:fewshot,lu:learn2explain} that constructed the CoT explanation by generating the reasoning after the answer, which has been proven to lead to better performance.

% As a superior method to elicit the reasoning ability of PLMs, Chain-of-Throught technique has been explored. 
% When the path is used before the final answer~\cite{suzgun:bigbench-cot,jason:cot,chen-2023-large,wang-iclr-2023-selfconsistency,zhou:least2most,li:reasoners}, it can improve the performance by decomposing a complex task.
% When the path is used after the final answer~\cite{stacey:humantouch,hase;when,marasovic:fewshot,mishra:instructions,lampinen:explanation,lu:learn2explain}, it can improve models' explanation, reliability, as well as the performance.
% \jlfu{Related works about CoT (you can make the works mentioned in intro more detailed.)}
% In this context, prompt tuning has become an essential technique to adapt large language models to specific tasks~\cite{liu2021pre,lester-etal-2021-power}. 

% In DST tasks, there is still no relevant work exploring the technique to introduce the chain-of-thought reasoning to assist state tracking. 
% \paragraph{Pre-trained Language Models}

% \jlfu{
% Paragraph1: related works about DST; 
% Paragraph3: Pre-trained Language Models (If we have enough space available.); 
% Paragraph2: related works about CoT (you can make the works mentioned in intro more detailed.)
% }

% \jlfu{@XuLin}

% \section{Chain-of-Thought Explanation for DST}
\section{Method}

\begin{figure*}[htbp]
    \centering
    \includegraphics[width=0.98\linewidth]{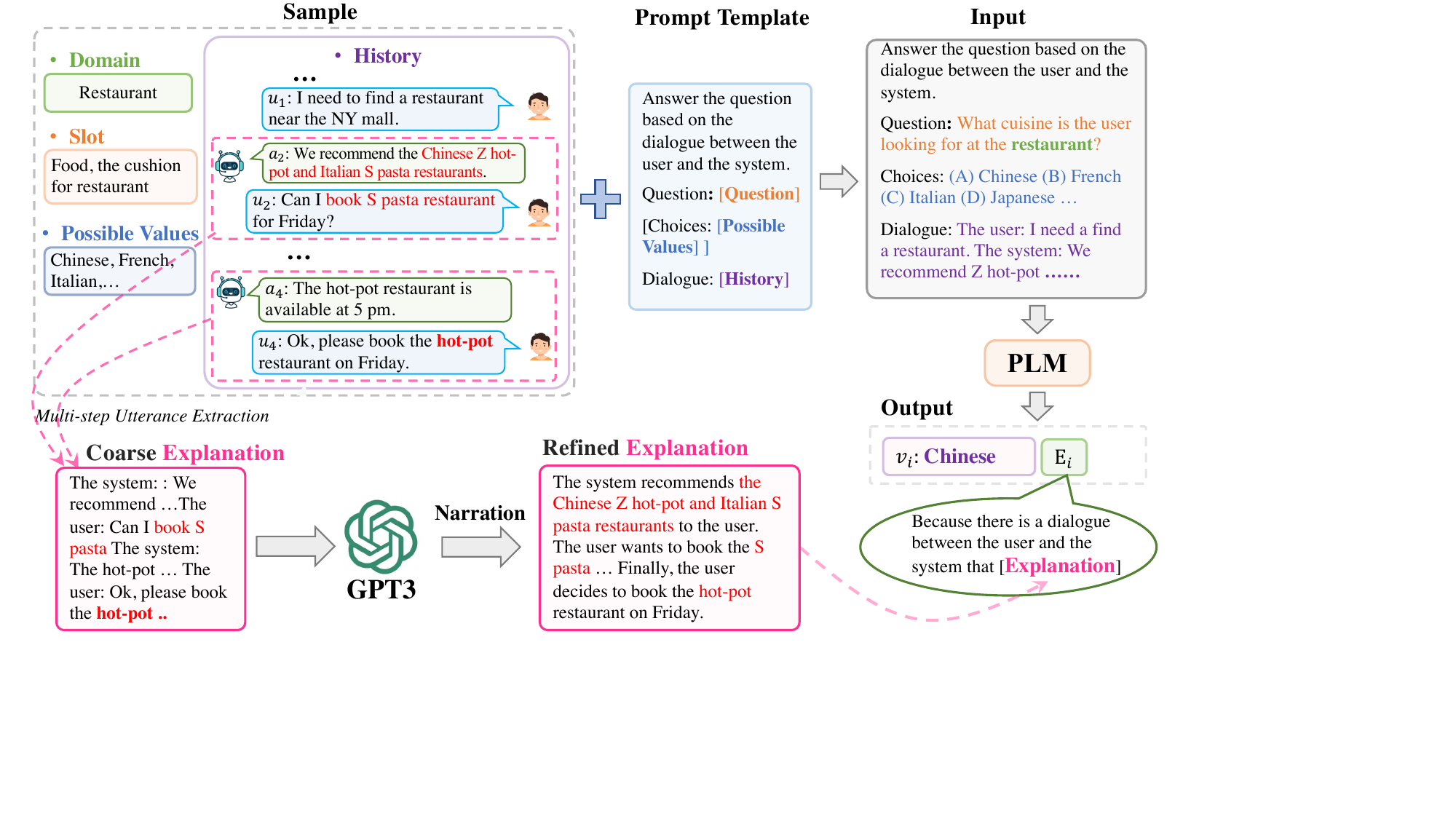}
    % \vspace{-7pt}
    \caption{The framework of CoTE. A DST sample is inserted to prompt templates to form inputs to PLMs. The outputs include both slot value $v_i$ and the Explanation $E_i$, which can be \textit{Coarse Explanation} or \textit{Refined Explanation}. 
    The former is the concatenation of dialogue snippets extracted to be used as the explanation, while the latter is the coarse explanation narrated by GPT3.
    % Utterance Explanations or the GPT3 narrations. The latter is rewritten from the former by GPT3.
    % \jlfu{The Output and the Sample are confusing. Why the output has two parts?}
    }
    \label{fig:framework}
\end{figure*}

\subsection{Task Definition}

Dialogue state tracking involves identifying and extracting essential attributes from conversations between a user and a system, with the purpose of fulfilling the user's goals. These attributes are represented as a collection of ``slot-value" pairs.  Each slot is usually defined with a schema according to its properties, formulated as a tuple $s_i=(\text{dm}_i, \text{sn}_i, \text{sd}_i, \text{pv}_i)$, where $\text{dm}_i$, $\text{sn}_i$ and $\text{sd}_i$ are the domain, name, and description of the slot. $\text{pv}_i$, as the possible value set, is available when the slot $s_i$ is a categorical slot. All these slots form the whole schema $\mathcal{S}=\{s_1,..,s_m\}$ for a DST task. The purpose of DST is to fill value to these slots according to the conversation history $\mathcal{D}$,  which comprises a series of system-user conversation turns, denoted as $\mathcal{D} = {(a_1, u_1), (a_2, u_2), ..., (a_i, u_i), ..., (a_N, u_N)}$. Here, $a_i$ and $u_i$ are the system and user utterances at time step $i$, and $N$ is the number of dialogue turns.

\subsection{Chain-of-Thought Explanation}
\label{sec:cote}
Sequence-to-sequence generation method with pretrained language models has demonstrated effectiveness and generalizability in completing many traditional NLP tasks~\cite{raman-etal-2022-transforming, saxena-etal-2022-sequence}, including DST~\cite{zhao-etal-2021-effective-sequence,lee-etal-2021-dialogue, cao-2022-d3st}. In these scenarios, prompting learning~\cite{liu2021pre} is commonly adopted. For example, in DST tasks, these methods leverage prompt templates to amalgamate various components such as dialogue history, slot domain, slot name, and slot description, which are pertinent to the target, into coherent natural language text. 
Subsequently, this text is utilized as input for the pretrained language model, with the slot values generated as the output.
\textbf{Notably, what distinguishes our CoTE from preceding prompt-based methods is its incorporation of explanations alongside slot values, thereby emphasizing the importance of the reasoning process.}

Formally, in a dialogue turn $t$, we design a prompt for a target slot with the dialogue context $\mathcal{D}_t$ and its relevant schema $s_i$. To incentivize the PLM to engage in reasoning when generating the slot values, we formulate the output as a composite of the slot's value, $v_i$, and its corresponding chain-of-thought explanation, denoted as $E_i$. During our experiments, we achieved better performance with the format of slot value followed by explanation, which can also largely reduce the generation overhead during inference by only generating the slot value. This procedure can be represented as follows:
\begin{equation}
v_i, E_i = \textsc{PLM}(\text{$\mathcal{T}$}(\mathcal{D}_t, s_i, dm_i, pv_i))
\end{equation}
where $dm_i$ is the domain name, $pv_i$ is the possible values of the slot $s_i$, and $\mathcal{T}$ is the prompt template, as shown in \autoref{fig:framework}.
PLM denotes the abbreviation of ``pre-trained language model''. 
$E_i$ is the explanation, which can be coarse or refined, as illustrated in \autoref{fig:framework}. Next, we will give a detailed introduction to the construction of coarse and refined explanations.

\paragraph{Coarse Explanation}
We propose a simple but effective way to construct coarse explanations for DST tasks. Specifically, relevant system-user utterance pairs that can contribute to changes in the slot values are extracted to form the explanation. Given the dialogue history $\mathcal{D} = [(a_1, u_1), (a_2, u_2), ..., (a_N, u_N)]$ and a requested slot, if the slot is applicable for the dialogue( has value in the dialogue history), there exists a subset of system-user utterance pairs $\hat{\mathcal{D}_s}\subset \mathcal{D}$ that are relevant the value of the requested slot. These utterance pairs might not be consecutive, as illustrated in Fig.~\ref{fig:framework}, where the example with $\hat{\mathcal{D}_s} = [(a_2, u_2), (a_4, u_4)]$. However,  \textbf{the sequential order among these utterance pairs can still reveal a certain narrative progression and logical relation. Therefore, we concatenate these extracted system-user utterances chronologically, forming a (multi-step) Chain-of-Thought explanation.} Since the explanations retain the dialogue format, we refer to it as the "coarse explanations."

\paragraph{Refined Explanation}
The above coarse explanations are constructed by concatenating the key sentences, which could lead to disfluencies, redundancies, or even grammar errors in the output text. To advance the effectiveness of CoTE, we seek ways to refine the fluency and interpretability of the generated explanation by converting the coarse explanations into narrative natural language. We call this as \emph{Refined Explanation}. Specifically, we use the GPT-3~\cite{gpt3-2020-nips} model from OpenAI without training to paraphrase the coarse explanations into a third-person narration, thus getting a revised explanation that is more fluent and explanation-like. We prompt the model with instructions and several paraphrased demonstrations to make GPT-3 acquire the ability to refine explanations. An example of GPT-3 third-person paraphrasing is shown in the top right of ~\autoref{fig:framework}. We use the refined explanations for training, as they are more semantically coherent and grammatically accurate. This refinement potentially facilitates improved correlation construction between slot values and their corresponding explanations, leading to enhanced performance in predicting slot values.

\paragraph{Prompt Template}
Our prompt template converts each DST sample into a question-answering format as shown in~\autoref{fig:framework}. For each data sample, we focus on generating the slot value for one target slot from the dialogue history and relevant slot schema. Given the \emph{Prompt Template}, we fill the corresponding placeholders, like \emph{[History]}, with the corresponding values to get the input to PLMs. \emph{[Question]} is constructed by automatically combining "What's" with the slot description. In case of incompatibility, we manually rewrite the question to make it semantically right; the questions are listed in our github repo~\footnote{\url{https://github.com/cathyxl/CoTE-DST}}. If the slot is categorical (with \emph{[Possible Values]}), then the question answer becomes a multiple-choice question (MCQ), thus attached with "Choices". For the non-categorical slot, we convert the prompt into an ordinary extractive QA question. 
% As for the "History" in the two templates, we concatenate all utterances with each led by its role phrase, "The user" or "The system", which is closer to natural compared to the special tokens used in ~\cite{lee-etal-2021-dialogue}.

\section{Experiment Settings}

\subsection{Datasets}
\begin{table}[h]
  \setlength{\belowcaptionskip}{-0.4cm}
  \renewcommand\tabcolsep{2.7pt}
  \centering
  \resizebox{0.9\linewidth}{!}{
  \begin{tabular}{lccc}
    \cmidrule[\heavyrulewidth]{1-4}
    \textbf{Datasets} & \textbf{Data Splits} & \textbf{\# Domains} & \textbf{\# Slots} \\
    \cmidrule{1-4}
    MultiWOZ 2.2 & 8445/1000/1000 & 8  & 49 \\
    M2M-R+M      & 1500/469/1039  & 2  & 14  \\
    WOZ 2.0      & 600/200/400    & 1  & 3  \\
    \cmidrule{1-4}
    \cmidrule[\heavyrulewidth]{1-4}
  \end{tabular}
  }
  % \vspace{-7pt}
  \caption{Datasets Statistics.}
  \label{table:datasets}
\end{table}
We conduct experiments on three widely-adopted datasets: MultiWOZ 2.2 \cite{budzianowski-etal-2018-multiwoz}, Machine Talking To Machines (M2M) \cite{loshchilovdecoupled}, and WOZ 2.0 \cite{mrksic-etal-2017-neural}. These datasets were chosen according to their distinct properties, enabling us to evaluate the ability to track dialogue states across various dialogue domains, literary styles, and complexities. The statistics are shown in ~\autoref{table:datasets}. 

\noindent\textbf{MultiWOZ 2.2} is a human-to-human, multidomain, task-oriented dialogue dataset featuring over 10,000 dialogues spanning 8 domains. MultiWOZ 2.2 provides a well-formatted schema for the DST task, including domain and slot descriptions and possible values for categorical slots.

\noindent\textbf{M2M} contains 3000 simulated dialogues from 2 domains, which are later refined by humans.  We refer to its sub-dataset from the restaurant and movie domains as M2M-R, and M2M-M respectively. The full version is denoted as M2M-R+M.

\noindent\textbf{WOZ 2.0} is built from a Wizard-Of-Oz (WOZ) experiment \cite{wen-etal-2017-network} where users type text in their queries instead of relying on speech inputs, thus have more complex dialogue contexts.

\subsection{Baselines}
We compare our CoTE-DST with most of the existing pretrained language model-based generative DST models, including the early pioneers \textbf{SimpleTOD}~\cite{hosseini2020simple} and \textbf{Seq2seq-DU}~\cite{feng2020seqseq}, as well as the further explorations employing larger PLMs: \textbf{AG-DST}~\cite{tian-etal-2021-amendable}, \textbf{SPACE-3}~\cite{he2022unified} and \textbf{TOATOD}~\cite{bang-etal-2023-task}. Additionally, we consider a bunch of schema-based methods akin to our setup, including \textbf{SDP}~\cite{lee-etal-2021-dialogue}, \textbf{DS2}~\cite{shin-etal-2022-dialogue} and \textbf{D3ST}~\cite{cao-2022-d3st}. Our CoTE has two variants: \textbf{CoTE-Coarse} and \textbf{CoTE-Refined}. The difference between them lies in whether the explanation is coarse or refined (by GPT3), as illustrated in Sec.~\ref{sec:cote}. We thoroughly investigated their capability in the experiments.

\subsection{Evaluation Metric}
We adopt the generally used Joint Goal Accuracy (JGA) as the evaluation metric. JGA measures the percentage of turns where the predicted values for all slots match the reference values exactly, we used the official evaluation script from the DSTC8 challenge\footnote{\url{https://github.com/google-research/google-research/tree/master/schema\_guided\_dst\#evaluation-on-multiwoz-21}}. The JGA score is highly suitable for the DST task, as it emphasizes the accumulated correctness of predicted slot values throughout each turn, which requires models to have strong reasoning abilities to closely track the variations of the user's intention as the dialogue evolves.

% Formally, let $D = \{d_1, d_2, ..., d_N\}$ be a dialogue with $N$ turns, $S = \{s_1, s_2, ..., s_N\}$ be the slot value in each turn, and $S' = \{s'_1, s'_2, ..., s'_N\}$ be the predicted values of slots in each turn. For each prediction $s'_t$, it is considered as correct if and only if the predictions and the ground truths match each other in the $t$-th turn as well as all the previous $t-1$ turns. JGA is calculated as $JGA = t_c/N$, where $t_c$ is the number of correct slot values in the whole dialogue. The JGA score is highly suitable for DST task, as it emphasizes the accumulated correctness of predicted slot values throughout each turn, which requires models to have strong reasoning abilities to closely track the variations of the user's intention as the dialogue evolves. During experiments, we used the official evaluation script from the DSTC8 challenge\footnote{\url{https://github.com/google-research/google-research/tree/master/schema\_guided\_dst\#evaluation-on-multiwoz-21}}. 

\subsection{Hyperparameters and Setup}

For the experiments, we employ the T5-base (220M) model and utilize the pretrained checkpoint from the Transformers\footnote{\url{https://huggingface.co/t5-base}} library. During the finetuning, we train the model using a batch size of 64 and a learning rate of 5e-5. Both models are trained with the AdamW optimizer~\cite{loshchilovdecoupled}. 

MultiWOZ 2.2, provides domain and slot descriptions, which are directly used in our CoTE construction. Following many previous works\cite{wu-etal-2019-transferable}, we also exclude the police and hospital domains. We adapted slot and domain descriptions from MultiWOZ 2.2 for datasets M2M and WOZ 2.0. We use this script\footnote{\url{https://github.com/google-research/google-research/blob/master/schema\_guided\_dst/multiwoz/create\_data\_from\_multiwoz.py}} to convert M2M and WOZ 2.0 into MultiWOZ 2.2 format. We will release all the processed data and source code.

% \subsection{Results}
\section{Experiment and Analysis}

\subsection{Exp-I: Effect of CoTE}
\paragraph{MultiWOZ 2.2}
According to ~\autoref{tab:main-exp-multiwoz}, our CoTE-Coarse has achieved decent performance compared to most of the baselines, except the most recent method TOATOD, which adopts T5-base initialized with more powerful weights and further refined with reinforcement learning.  Notably, CoTE-Coarse has outperformed all of the schema-based baselines based on the same pretrained language model T5$_\text{base}$ as we used, SDP and D3ST, even better than the model DS2 that exploits T5-large. The improved performance of CoTE-Coarse validates the effectiveness of the proposed chain-of-thought explanation technique, indicating that incorporating explanations can lead to more accurate slot values. Furthermore, the enhanced CoTE-Refined model achieves better Joint Goal Accuracy (JGA) than CoTE-Coarse. This proves that providing more natural and compatible explanations following the slot value answer can assist the model in capturing slot values more accurately.

\begin{table}[t]\small
    \centering
    \begin{threeparttable}
    \resizebox{\linewidth}{!}{
    \begin{tabular}{l c c c}
        \toprule
        \multirow{3}{*}{\textbf{Models}} & \multirow{3}{*}{\textbf{No. Param}}  & \multicolumn{2}{c}{\textbf{Dataset}} \\
        \cmidrule{3-4} &  & MWZ 2.2  & WOZ 2.0 \\
        % \midrule
        % BERT-DST \cite{chao2019bert} & - & 87.7 \\
        % SGD-baseline~\cite{rastogi2020towards} & 42.0 &- \\
        % TripPY~\cite{heck-etal-2020-trippy} & 53.5 & -\\
        % DS-DST~\cite{zhang-etal-2020-find} & 51.7 & - \\
        \midrule
        % COMER \cite{ren-etal-2019-scalable} & -& 88.6 \\
        % TRADE~\cite{wu-etal-2019-transferable} & 48.6 & -\\
        Seq2seq-DU & BERT(100M)         & 54.4  & 91.2\\
        SimpleTOD  & DistillGPT2(82M)  & 54.0$^\dag$ & - \\
        AG-DST     & Plato2(310M)      & 57.3 & 91.4 \\
        SPACE-3    & UniLM(340M)        & 57.5  &  - \\
        TOATOD     & T5$_{\text{base}}$(220M)$^\ast$ & 63.8 & - \\
        % DICOS*~\cite{guo-etal-2022-beyond} & 40.3 & 86.0 \\
        SDP       & T5$_{\text{base}}$(220M)       & 56.4$^\diamond$  & 86.6\\
        DS2      & T5$_{\text{large}}$(770M)       & 52.1$^\star$ & \textbf{92.5}\\
        D3ST       & T5$_{\text{base}}$(220M)      & 56.1 & - \\
        \midrule
        CoTE-Coarse & T5$_{\text{base}}$(220M)     & 57.0 & 88.2 \\
        CoTE-Refined & T5$_{\text{base}}$(220M)    & \textbf{57.5} & 91.6 \\
        \bottomrule
    \end{tabular}
    }
    \end{threeparttable}
    \vspace{-7pt}
    \caption{The Joint Goal Accuracy (JGA) on MultiWOZ(MWZ) 2.2 and WOZ 2.0. 
    % '$\ddag$': our reproduction results by official code. 
    `$\dag$': borrowed from AG-DST~\cite{tian-etal-2021-amendable}. `$\ast$': TOATOD is built on a frozen T5$_{\text{base}}$ model initialized with pretrained weights from PPTOD\cite{su-etal-2022-multi}, which is trained on a large dialogue dataset. It is then finetuned with a DST adapter(36M) and is further refined using reinforcement learning.
    `$\diamond$': This result was obtained using the official code base of SDP, and our CoTE (a model based on SDP) also uses the same code base to ensure a fair comparison.
    `$\star$': We run the official code of DS2 to get the result because it lacks MultiWOZ(MWZ) 2.2 results.
    }
    \label{tab:main-exp-multiwoz}
\end{table}

\paragraph{WOZ 2.0}
Comparable findings can be observed in an alternative single-domain dialogue state tracking (DST) dataset called WOZ 2.0, as presented in ~\autoref{tab:main-exp-multiwoz}. Our CoTE-Coarse model surpasses the majority of other baselines, and CoTE-Refined exhibits superior outcomes compared to CoTE-Coarse. 
In this table, DS2 employs a T5-large model that was pretrained on the SAMSum corpus~\cite{gliwa-etal-2019-samsum}, a dataset for text summarization. Consequently, DS2 achieves improved Joint Goal Accuracy (JGA) when its code is executed on the WOZ 2.0 dataset. On the other hand, our CoTE model is based on T5-base and does not possess any specialized pretraining.

\paragraph{M2M}

Within the M2M datasets~\autoref{tab:main-exp-m2m}, we conduct comparable analyses and observe that our models, CoTE-Coarse and CoTE-Refined, achieve superior or highly competitive outcomes in comparison to strong baselines like Seq2Seq-DU, SDP. Particularly, CoTE-Refined demonstrates improved performance on M2M-M and M2M-R+M, indicating that GPT3 narrative explanations offer assistance in handling more intricate datasets. This is supported by the fact that M2M-M generally exhibits lower Joint Goal Accuracy (JGA) while M2M-R+M encompasses a wider range of domains and slots.

\begin{table}[t]
    \centering \footnotesize
    \renewcommand\tabcolsep{3pt}
    % \resizebox{0.8\linewidth}{!}{
    \begin{tabular}{l c c c c}
        \toprule
        \multirow{3}{*}{\textbf{Models}} & \multirow{3}{*}{\textbf{No.Param}} &\multicolumn{3}{c}{\textbf{Dataset}} \\
        \cmidrule{3-5}
         & & R & M & R+M \\
        \midrule
        % BERT-DST~\cite{chao2019bert} & 89.6 & 80.1 & 86.9 \\
        % TripPY~\cite{heck-etal-2020-trippy} & 90.0 & 83.5 & - \\
        % \midrule
        Seq2seq-DU & BERT(110M) & - & - & \textbf{90.9} \\
        SDP &  T5$_{\text{base}}$(220M)  & 90.6 & 81 & 86.4 \\
        DS2 &   T5$_{\text{large}}$(770M) &  87.6$^\ddag$ & 70.9$^\ddag$ & 74.6$^\ddag$ \\
        % DICOS*~\cite{guo-etal-2022-beyond} & 91.5 & 84.7 & 82.8 \\ 
        \midrule
        CoTE-Coarse & T5$_{\text{base}}$(220M) &\textbf{91.8} & 85.1 & 88.4 \\
        CoTE-Refined & T5$_{\text{base}}$(220M) & 90.3 & \textbf{85.3} & 90 \\
        \bottomrule
    \end{tabular}
    \vspace{-7pt}
    \caption{The Joint Goal Accuracy (JGA) on M2M-R (R), M2M-M (M), and M2M-R+M (R+M).'$\ddag$': our reproduction results by official code.
    % \vspace{-3pt}
    }
    
    \label{tab:main-exp-m2m}
\end{table}

\subsection{Exp-II: Effects of CoTE in Low-resource}
In addition to conducting experiments on the full dataset, we delve into the low-resource setting, where only a fraction of the training data is used during model training. Evaluating performance in this experiment setting provides deeper insights into the model's effectiveness in utilizing limited data samples and its generalization capabilities.
\subsubsection{Settings}
We choose to train the model using randomly sampled training data, comprising 1\%, 5\%, 10\%, and 20\% of the training data samples. To mitigate potential instability stemming from limited training samples, we conduct multiple low-resource trainings. The results are then averaged across three randomly generated seeds for the MultiWOZ 2.2 dataset and across five seeds for the M2M-R+M and WOZ 2.0 datasets.

\subsubsection{Results}
The main observations are summarized below:

\paragraph{O1: CoTE variants outperform competitive baselines in low-resource settings.} 
% Even when some methods outperform CoTE variants in a full-data setting, they are exceeded by CoTE variants in the low-shot setting. For example, as shown in~\autoref{tab:main-exp-multiwoz}, DS2's JGA 92.5 outperforms CoTE-Refined's 91.6 on full-data WOZ 2.0, but as shown in~\autoref{tab:low-shot}, CoTE-Refined largely exceeded DS2 in the 5\%,10\%, and 20\% low-shot settings, which indicates training chain-of-thought explanation is more capable and efficient in eliciting the model's reasoning ability, thus having much better generalization ability than using the summarization method. When compared to SDP, CoTE variants also achieve better result, which can directly prove the generalization ability of CoTE. However, we noticed in the experiments on dataset WOZ 2.0, that DS2 shows a better result in the 1\% setting. We explain the results as follows: DS2 is finetuned based on a T5-large model pre-trained on summarization task, which could perform better due to its more powerful parameters than our backbone T5-base.
While certain methods may demonstrate superior performance over CoTE variants with full data, these roles reverse dramatically in low-resource conditions. For instance, in a full-data scenario, DS2's JGA of 92.5 surpasses CoTE-Refined's 91.6 on WOZ 2.0 (as illustrated in~\autoref{tab:main-exp-multiwoz}). However, when examining low-resource conditions (as depicted in~\autoref{tab:low-shot}), CoTE-Refined significantly outperforms DS2 across 5\%, 10\%, and 20\% settings. This suggests that training with a chain-of-thought explanation method is remarkably effective at eliciting the model's reasoning abilities because it emphasizes more about the deducing process to get the slot value instead of the overall summarization, thereby enhancing its generalization with even smaller data samples. Furthermore, when compared to SDP, CoTE variants consistently yield superior results, underscoring their enhanced generalization capabilities. Nonetheless, it's noteworthy that DS2 exhibits better performance in the 1\% setting during experiments on the WOZ 2.0 dataset. We attribute this result to DS2's finetuning on the larger T5-large model pretrained for summarization tasks, leveraging its robust parameters, which outmatch those of our T5-base backbone model.
% \paragraph{O1: CoTE variants outperform competitive baselines in low-shot settings.} Even when some methods outperform CoTE variants in a full-data setting, they are exceeded by CoTE variants in the low-shot setting. For example, as shown in~\autoref{tab:main-exp-multiwoz}, DS2's JGA 92.5 outperforms CoTE-Refined's 91.6 on full-data WOZ 2.0, but as shown in~\autoref{tab:low-shot}, CoTE-Refined largely exceeded DS2 in the 5\%,10\%, and 20\% low-shot settings, which indicates training chain-of-thought explanation is more capable and efficient in eliciting the model's reasoning ability. We also noticed in the experiments on dataset WOZ 2.0, that DICOS and DS2 show better results in the 1\% setting. We explain the results as follows: While running low resource experiments for DICOS on WOZ 2.0, we used the gold update indicator because the original paper didn't provide the learned update predictor, so the result is above the level of the normal DICOS. DS2 is finetuned based on a T5-large model pre-trained on summarization task, which could perform better due to its more parameters than our backbone T5-base.
\paragraph{O2: CoTE variants show larger improvement margins when sparse data samples.} As shown in~\autoref{tab:low-shot}, comparing SDP's results with our CoTE-Coasrse on MultiWOZ 2.2, the margins for low-resource 1\%, 5\%, 10\% 20\% are 2, 1.1, 0.2, 0.4, respectively. These indicate that our CoTE can withstand data sparsity; further validating CoTE strengthens the model's generalizability. 
\paragraph{O3: CoTE-Refined outperforms CoTE-Coarse more in scenarios of fewer data sample.} In~\autoref{tab:low-shot}, the JGA difference between CoTE-Refined and CoTE-Coarse on settings 1\%/5\%/10\%/20\%, are 0.5/0.5/0.2/0.4, 6.4/9.9/2.2/0, 0.3/2.1/6.3/3 for MultiWOZ 2.2, M2M-R+M and WOZ 2.0, respectively, showing decreasing tendency. For WOZ2.0, since the data samples are too sparse for 1\%(4 training dialogues) and 5\%, we can still observe reducing margins from 10\% to 20\%. From this observation, GPT3 refined explanations could be more effective on suitably fewer data samples.

\begin{table*}[!htb]
    \renewcommand\tabcolsep{2.7pt}
    % \centering \footnotesize
    \resizebox{\linewidth}{!}{
    \begin{tabular}{llllllllllllllll}
        \toprule
        \multirow{2}{*}{\textbf{Methods}}
        % \cmidrule{2-2} 
        & \multicolumn{5}{c}{MultiWOZ 2.2} & \multicolumn{5}{c}{M2M-R+M} & \multicolumn{5}{c}{WOZ 2.0}  \\
        \cmidrule{2-16} 
        % \cmidrule(lr)(2-5)\cmidrule(lr)(6-9)\cmidrule(lr)(10-13)
        & 1\% & 5\% & 10\% & 20\% & 100\% & 1\% & 5\% & 10\% & 20\% & 100\% & 1\% & 5\% & 10\% & 20\% & 100\% \\
        \midrule
        % DICOS  & 20.7$^{\dag,\ddag}$ & 35.6$^{\dag,\ddag}$ & 37.4$^{\dag,\ddag}$ & 37.8$^{\dag,\ddag}$ & 40.4$^{\dag,\ddag}$   & 20.0$^{\dag,\ddag}$ & 53.9$^{\dag,\ddag}$ & 69.4$^{\dag,\ddag}$ & 77.6$^{\dag,\ddag}$ & 82.8$^{\dag,\ddag}$   & \textbf{34.9} & 60.5$^{\dag,\ddag}$ & 68.4$^{\dag,\ddag}$ & 70.3$^{\dag,\ddag}$ & 86.0$^{\dag,\ddag}$ \\
        DS2    & 34.8$^{\dag,\ddag}$ & 45.1$^{\dag,\ddag}$ & 47.7$^{\dag,\ddag}$ & 49.3$^{\dag,\ddag}$ & 52.1$^{\dag,\ddag}$   & 38.3$^{\dag,\ddag}$ & 62.3$^{\dag,\ddag}$ & 67.5$^{\dag,\ddag}$ & 72.8$^{\dag,\ddag}$ & 74.6$^{\dag,\ddag}$   & 32.1 & 63.2$^{\dag,\ddag}$ & 75.9$^{\ddag}$ & 83.7$^{\ddag}$ & \textbf{92.5}\\
        SDP    & 38.1$^{\dag,\ddag}$ & 49.4$^{\dag,\ddag}$ & 53.3$^{\dag}$ & 54.7$^{\ddag}$ & 56.4$^{\dag,\ddag}$   & 33.1$^{\dag,\ddag}$  & 55.7$^{\dag,\ddag}$  & 59.7$^{\dag,\ddag}$  & 63.1$^{\dag,\ddag}$  & 86.4$^{\dag,\ddag}$    & 12.1$^{\dag,\ddag}$  & 53.1$^{\dag,\ddag}$  & 64.3$^{\dag,\ddag}$  & 74.5$^{\dag,\ddag}$  & 86.6$^{\dag,\ddag}$ \\
        \midrule
        CoTE-Coarse   & 40.1$^{\ddag}$ & 50.5$^{\ddag}$ & \textbf{54.4} & 55.0 & 57.0$^{\ddag}$   & 42.1$^{\ddag}$ & 63.4$^{\ddag}$ & 77.8$^{\ddag}$ & 83.4$^{\ddag}$ & 88.4$^{\ddag}$   & 22.3 & 62.8$^{\ddag}$ & 72$^{\ddag}$ & 81.8$^{\ddag}$ & 88.2$^{\ddag}$ \\
        CoTE-Refined & \textbf{40.6} & \textbf{51.0} & 53.7 & \textbf{55.4} & \textbf{57.5}    & \textbf{48.5} & \textbf{73.3} & \textbf{80} & \textbf{83.4} & \textbf{90}    & 22.6 & \textbf{64.9} & \textbf{78.3} & \textbf{84.8} & 91.6\\
        \bottomrule
    \end{tabular}
    }
    % \vspace{-7pt}
    \caption{Low resource comparsions on MultiWOZ 2.2,  M2M-R+M and WOZ 2.0.  
    $\dag$ and $\ddag$ denote that CoTE-Coarse and CoTE-Refined significantly outperform the baseline method, respectively.
    % We've also done significance tests between SDP and our CoTE-refined, the p-values for 1\% and 5\% are all lower than 0.25 and  showing that
    % \jlfu{1. We should add a significance test for CoTE and CoTE-refined to show our methods (CoTE and CoTE-refined ) is significantly better than the baselines model in different settings. 2. Can we add the performance of the full dataset?}
    }
    \label{tab:low-shot}
\end{table*}

\subsection{Exp-III: Where Does CoTE Work?}

\subsubsection{Fine-grained Evaluations}
% \paragraph{Aspect Definition}
Dialogue state tracking task often presents challenges due to complex dialogue contexts, such as longer turns or utterances, which indicates more noisy information and slots. Another difficulty comes from the multiple value alterations of a single slot in non-adjacent turns. To probe how different methods perform in the above scenarios, we design 3 fine-grained evaluation aspects. (1) \emph{Reasoning step}: \textbf{$\phi(\text{step})$}, the total number of turns in a dialogue, (2) \emph{Dialogue turns}: \textbf{$\phi(\text{turn})$}, the maximum number of slot value changes in one dialogue, and (3) \emph{Average utterance length} \textbf{$\psi(\text{len})$}, the average utterance length in a dialogue. Specifically, we split the testing data samples according to different ranges of the above three aspects. For \textbf{$\phi(\text{turn})$}, we choose spans 0-11, 11-14, 14-19, 20+ for MultiWOZ 2.2/M2M-R-M and 0-5, 6-7, 8-9, and 10+ for WOZ 2.0. For \textbf{$\phi(\text{step})$}, we split data samples with the number of reasoning steps falling in 1, 2, and 3+ for all three datasets. As for \textbf{$\psi(\text{len})$}, three ranges are divided differently for each dataset, 0-12, 12-15, 15+ for MultiWOZ 2.2, 0-8, 8-9, 10+ for M2M-R-M and 0-5, 6-7, 8+ for WOZ 2.0. We provide details about these ranges in our anonymous GitHub repo~\footnote{\url{https://anonymous.4open.science/r/cote-release-1847/README.md}}.

\subsubsection{Results}

\begin{table*}[h!]
  \centering \scriptsize
  \renewcommand\tabcolsep{0.3pt}
    \renewcommand\arraystretch{1} 
    \resizebox{\linewidth}{!}{
    \begin{tabular}{ cccc cccc cccc cccc cccc cccc cccc cccc cccc cccccc}
    \toprule
       \multicolumn{12}{c}{\textbf{Reasoning Step} \textbf{$\phi(\text{step})$}} &
        \multicolumn{16}{c}{\textbf{Dialog Turns} \textbf{$\phi(\text{turn})$}} &
        \multicolumn{14}{c}{\textbf{Average Utterance Length} \textbf{$\psi(\text{len})$}} \\
 % \midrule
 
 \cmidrule(lr){1-12}\cmidrule(lr){13-28}\cmidrule(lr){29-42}
 \multicolumn{42}{c}{\textbf{MultiWOZ 2.2}} \\
 \multicolumn{4}{c}{step=1} & 
 \multicolumn{4}{c}{step=2} & 
 \multicolumn{4}{c}{step=3+} & 
 \multicolumn{4}{c}{\# turn=0-9} & 
 \multicolumn{4}{c}{\# turn=10-14} & 
 \multicolumn{4}{c}{\# turn=15-19} & 
 \multicolumn{4}{c}{\# turn=20+} & 
 \multicolumn{4}{c}{len=0-11} & 
 \multicolumn{4}{c}{len=12-14} & 
 \multicolumn{6}{c}{len=15+}\\
    \midrule
\multicolumn{4}{c}{\multirow{5}[2]{*}{\includegraphics[width=0.095\linewidth]{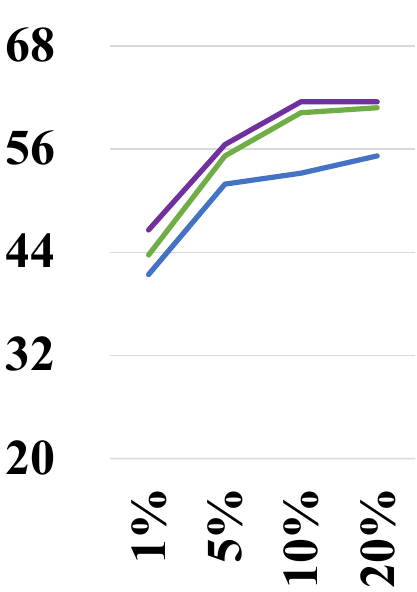}}} & 
\multicolumn{4}{c}{\multirow{5}[2]{*}{\includegraphics[width=0.095\linewidth]{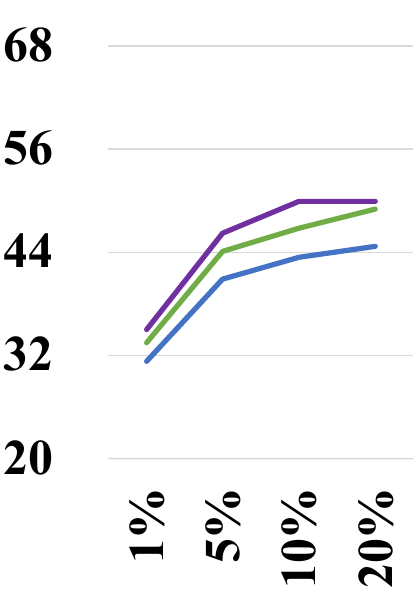}}} &
\multicolumn{4}{c}{\multirow{5}[2]{*}{\includegraphics[width=0.095\linewidth]{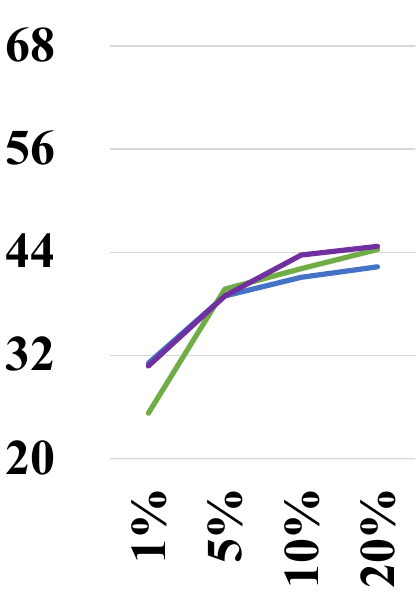}}} & 
\multicolumn{4}{c}{\multirow{5}[2]{*}{\includegraphics[width=0.095\linewidth]{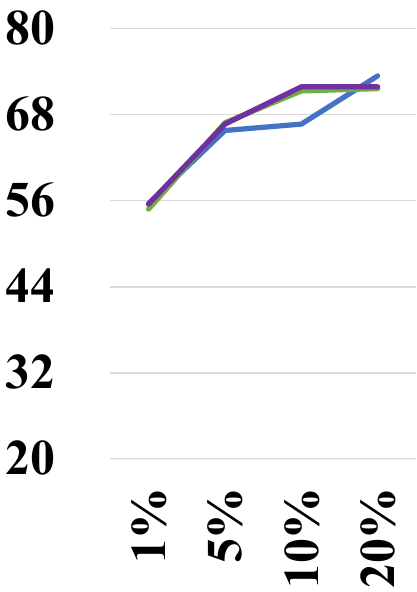}}} &
\multicolumn{4}{c}{\multirow{5}[2]{*}{\includegraphics[width=0.095\linewidth]{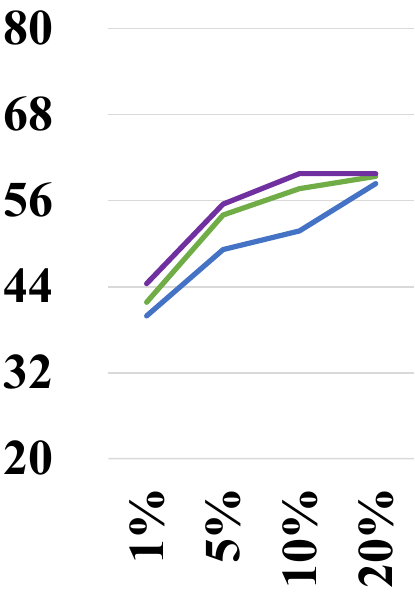}}} & 
\multicolumn{4}{c}{\multirow{5}[2]{*}{\includegraphics[width=0.095\linewidth]{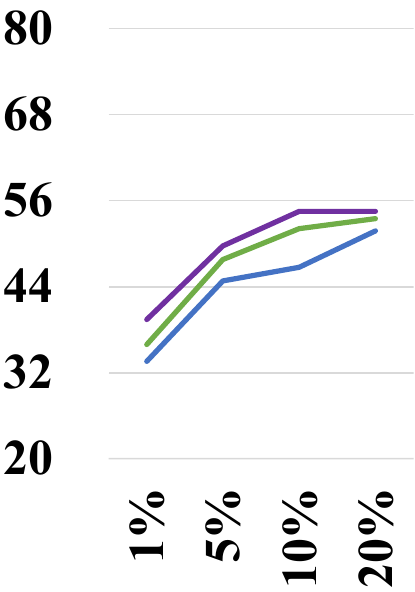}}} &
\multicolumn{4}{c}{\multirow{5}[2]{*}{\includegraphics[width=0.095\linewidth]{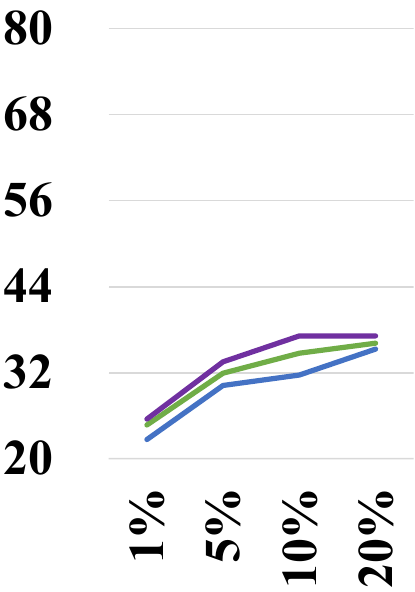}}} &
\multicolumn{4}{c}{\multirow{5}[2]{*}{\includegraphics[width=0.095\linewidth]{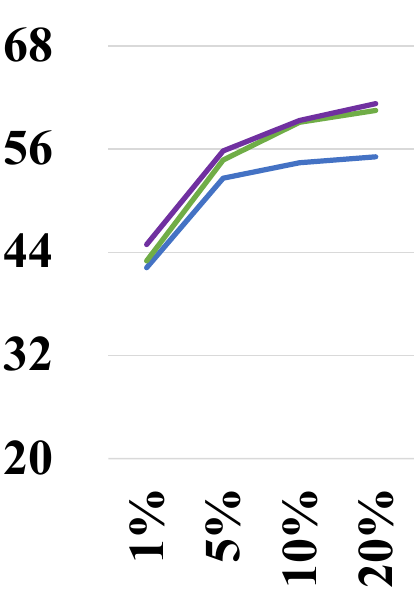}}} &
\multicolumn{4}{c}{\multirow{5}[2]{*}{\includegraphics[width=0.095\linewidth]{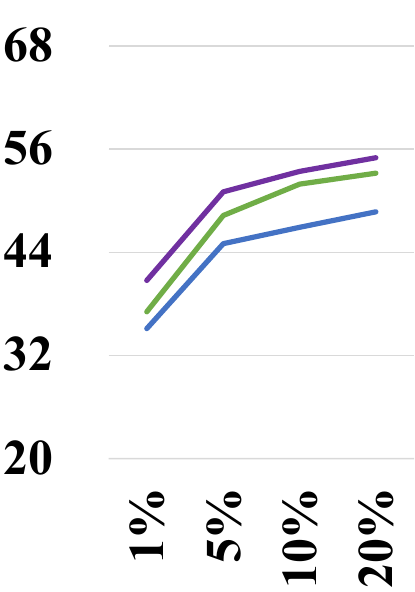}}} &
\multicolumn{4}{c}{\multirow{5}[2]{*}{\includegraphics[width=0.095\linewidth]{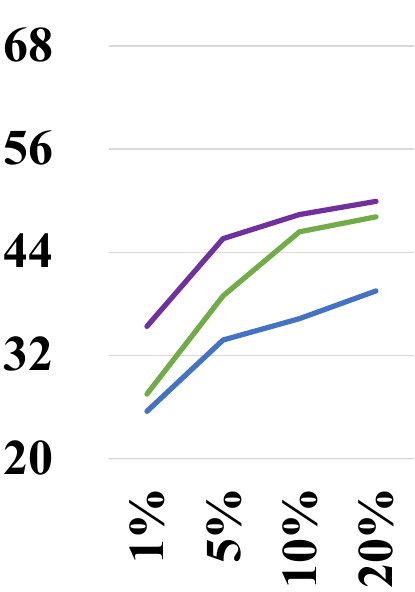}}} &
\multicolumn{2}{c}{\multirow{5}[2]{*}{\includegraphics[width=0.05\linewidth]{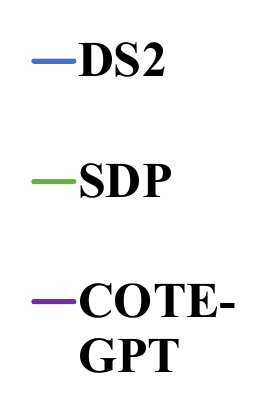}}} 

\\ \\ \\ \\ \\ \\ \\ \\
 \midrule
 \multicolumn{42}{c}{\textbf{M2M-R+M}} \\
\multicolumn{4}{c}{step=1} & 
 \multicolumn{4}{c}{step=2} & 
 \multicolumn{4}{c}{step=3+} & 
 \multicolumn{4}{c}{\# turn=0-9} & 
 \multicolumn{4}{c}{\# turn=10-14} & 
 \multicolumn{4}{c}{\# turn=15-19} & 
 \multicolumn{4}{c}{\# turn=20+} & 
 \multicolumn{4}{c}{len=0-7} & 
 \multicolumn{4}{c}{len=8-9} & 
 \multicolumn{6}{c}{len=10+} \\
 \midrule
\multicolumn{4}{c}{\multirow{5}[2]{*}{\includegraphics[width=0.095\linewidth]{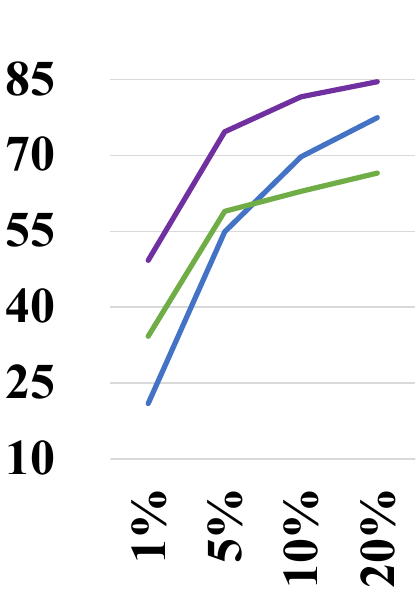}}} & 
\multicolumn{4}{c}{\multirow{5}[2]{*}{\includegraphics[width=0.095\linewidth]{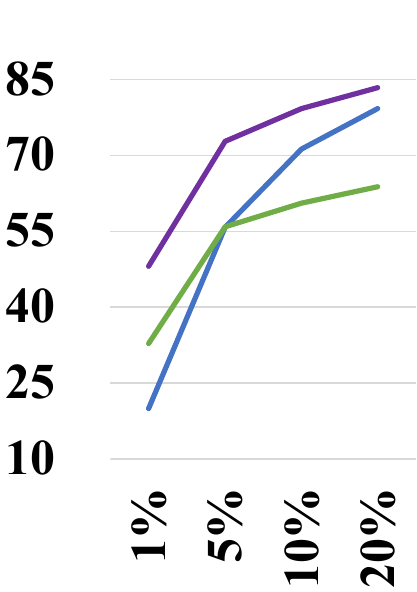}}} &
\multicolumn{4}{c}{\multirow{5}[2]{*}{\includegraphics[width=0.095\linewidth]{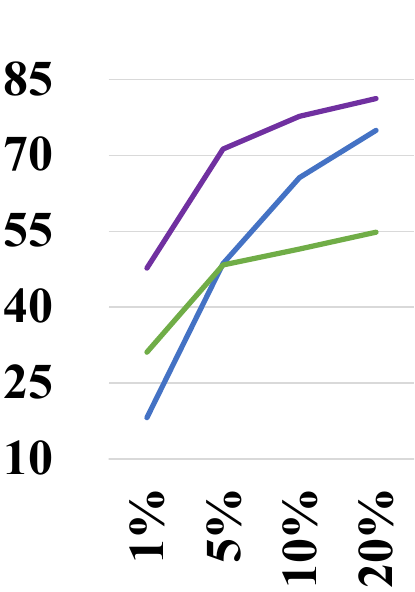}}} & 
\multicolumn{4}{c}{\multirow{5}[2]{*}{\includegraphics[width=0.095\linewidth]{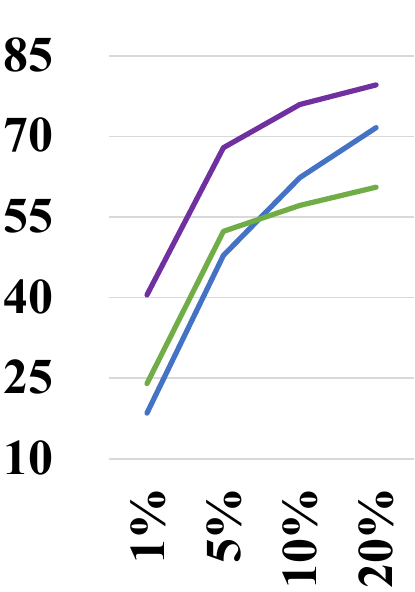}}} &
\multicolumn{4}{c}{\multirow{5}[2]{*}{\includegraphics[width=0.095\linewidth]{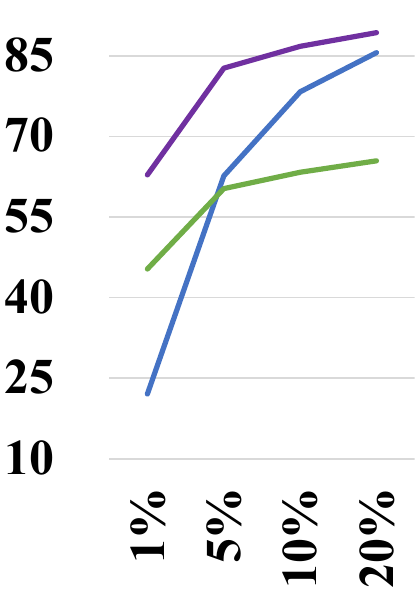}}} & 
\multicolumn{4}{c}{\multirow{5}[2]{*}{\includegraphics[width=0.095\linewidth]{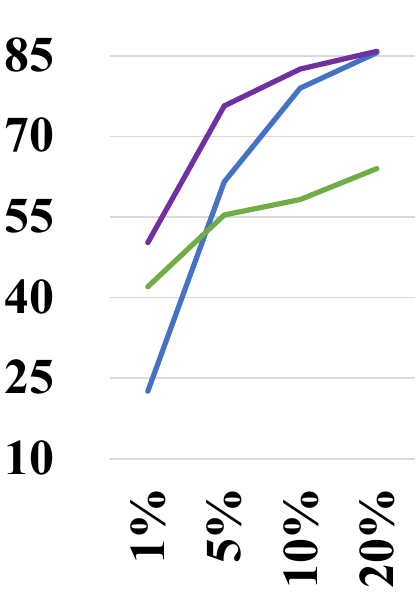}}} &
\multicolumn{4}{c}{\multirow{5}[2]{*}{\includegraphics[width=0.095\linewidth]{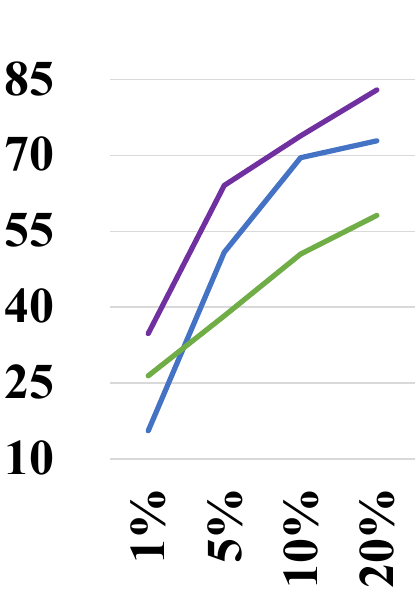}}} &
\multicolumn{4}{c}{\multirow{5}[2]{*}{\includegraphics[width=0.095\linewidth]{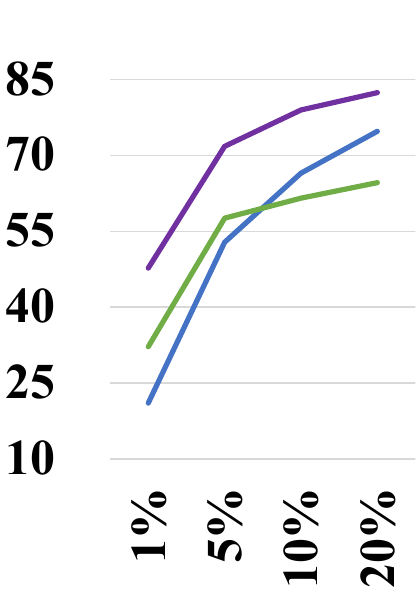}}} &
\multicolumn{4}{c}{\multirow{5}[2]{*}{\includegraphics[width=0.095\linewidth]{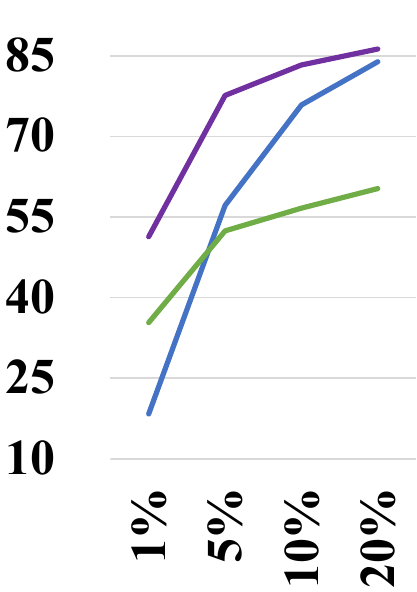}}} &
\multicolumn{4}{c}{\multirow{5}[2]{*}{\includegraphics[width=0.095\linewidth]{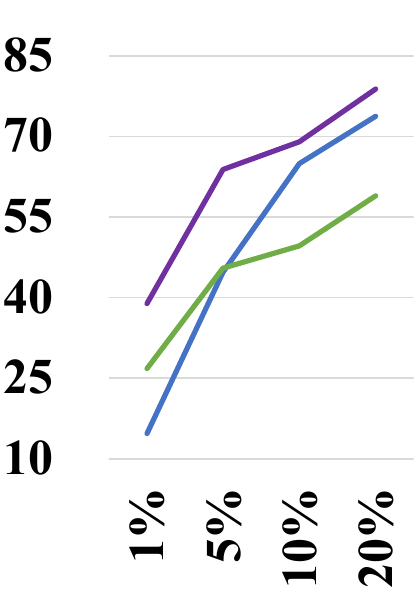}}} &
\multicolumn{2}{c}{\multirow{5}[2]{*}{\includegraphics[width=0.05\linewidth]{fig/fine_3m_line/legend.pdf}}} 
\\ \\ \\ \\ \\ \\ \\ \\ 
 \midrule
 \multicolumn{42}{c}{\textbf{WOZ 2.0}} \\
\multicolumn{4}{c}{step=1} & 
 \multicolumn{4}{c}{step=2} & 
 \multicolumn{4}{c}{step=3+} & 
 \multicolumn{4}{c}{\# turn=0-5} & 
 \multicolumn{4}{c}{\# turn=6-7} & 
 \multicolumn{4}{c}{\# turn=8-9} & 
 \multicolumn{4}{c}{\# turn=10+} & 
 \multicolumn{4}{c}{len=0-5} & 
 \multicolumn{4}{c}{len=6-7} & 
 \multicolumn{6}{c}{len=8+} \\
 \midrule
\multicolumn{4}{c}{\multirow{5}[2]{*}{\includegraphics[width=0.095\linewidth]{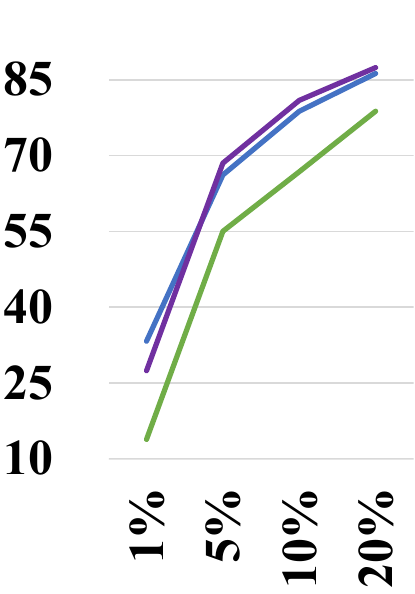}}} & 
\multicolumn{4}{c}{\multirow{5}[2]{*}{\includegraphics[width=0.095\linewidth]{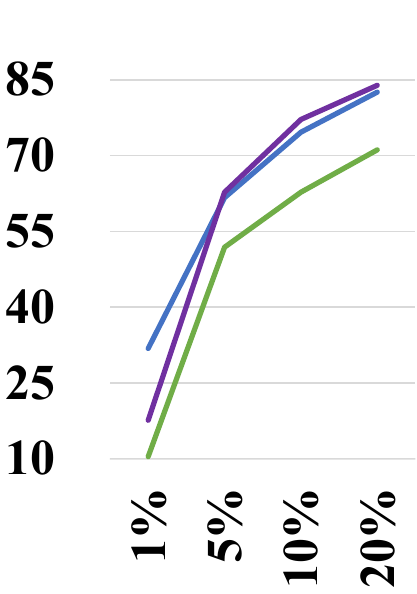}}} &
\multicolumn{4}{c}{\multirow{5}[2]{*}{\includegraphics[width=0.095\linewidth]{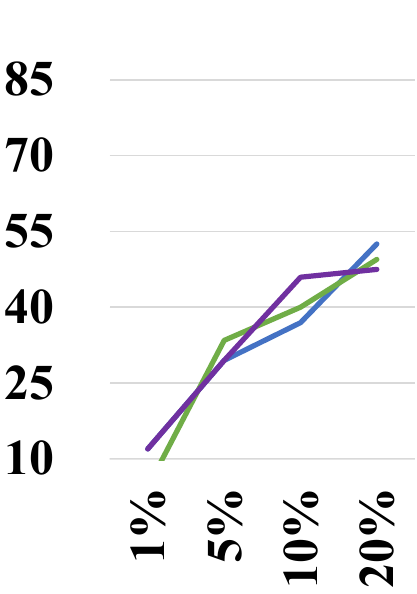}}} & 
\multicolumn{4}{c}{\multirow{5}[2]{*}{\includegraphics[width=0.095\linewidth]{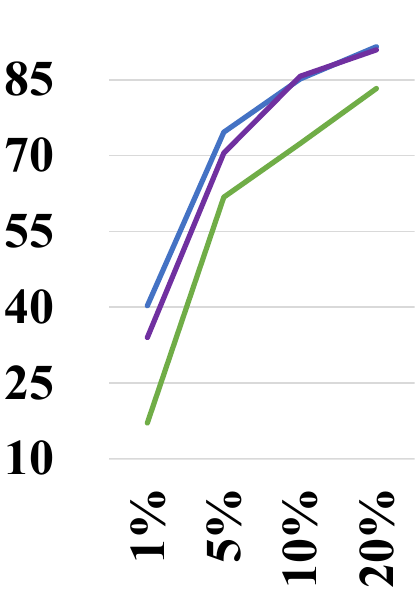}}} &
\multicolumn{4}{c}{\multirow{5}[2]{*}{\includegraphics[width=0.095\linewidth]{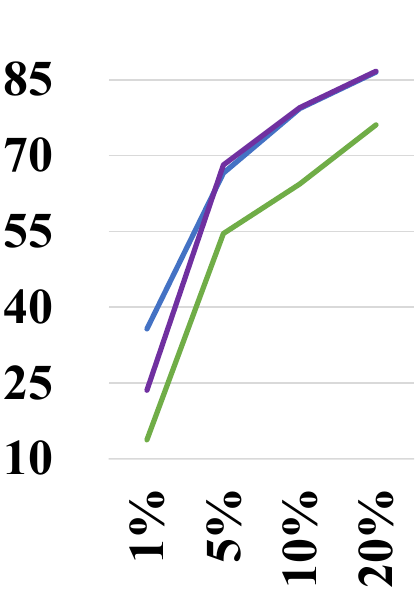}}} & 
\multicolumn{4}{c}{\multirow{5}[2]{*}{\includegraphics[width=0.095\linewidth]{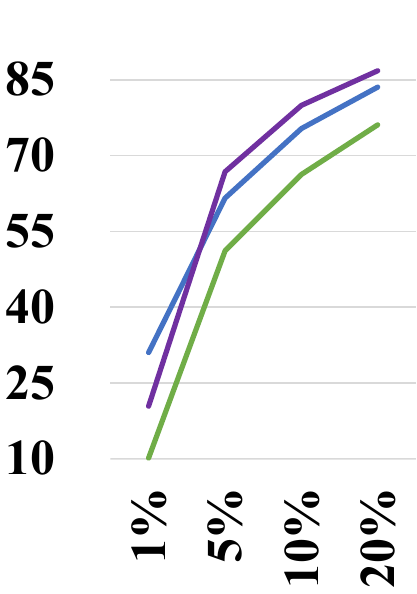}}} &
\multicolumn{4}{c}{\multirow{5}[2]{*}{\includegraphics[width=0.095\linewidth]{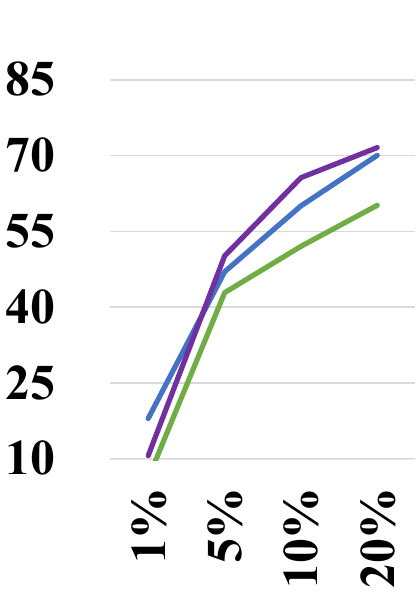}}} &
\multicolumn{4}{c}{\multirow{5}[2]{*}{\includegraphics[width=0.095\linewidth]{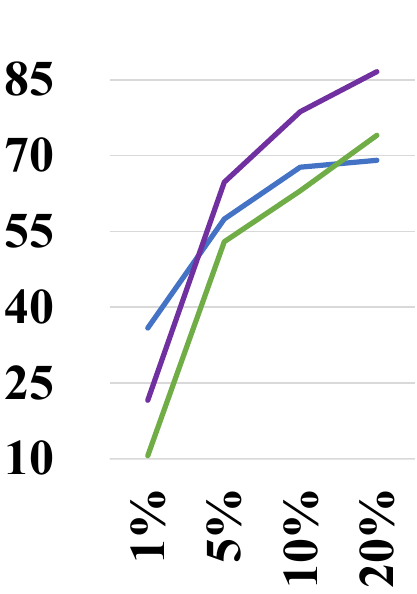}}} &
\multicolumn{4}{c}{\multirow{5}[2]{*}{\includegraphics[width=0.095\linewidth]{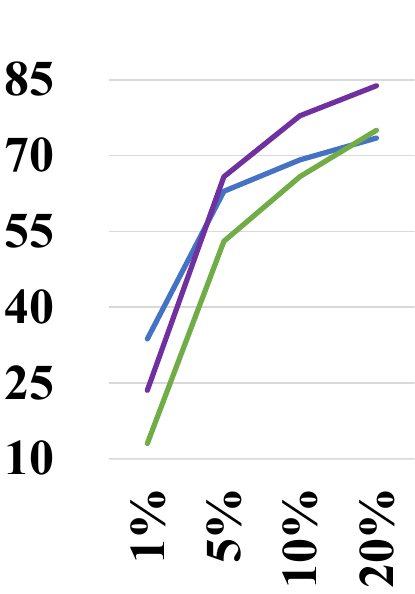}}} &
\multicolumn{4}{c}{\multirow{5}[2]{*}{\includegraphics[width=0.095\linewidth]{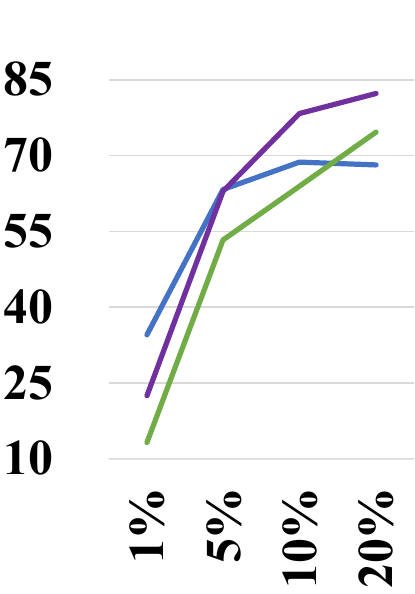}}} &
\multicolumn{2}{c}{\multirow{5}[2]{*}{\includegraphics[width=0.05\linewidth]{fig/fine_3m_line/legend.pdf}}} 
\\ \\ \\ \\ \\ \\ \\ \\ 
    \bottomrule 
    \end{tabular}
    }
    \vspace{-7pt}
      \caption{Fine-grained comparisons under low resource setting.}
      \vspace{-7pt}
  \label{tab:fine-grain}
\end{table*}
Fine-grained results are presented in ~\autoref{tab:fine-grain}. we observe the following results:
\paragraph{O1: Existing methods have difficulty dealing with complex examples.}
We assume that the longer the reasoning steps, the dialogue turns, or utterances, the greater the difficulty on DST. This assumption can be supported by~\autoref{tab:fine-grain}, where the line charts on the right-hand side in each fine-grained aspect indicate the lower lines. 

\paragraph{O2: CoTE variants lead the performance in most line charts, particularly in more complex samples.}
Notably, our methods present large improvement margins in dialogues where the reasoning steps, the number of dialogue turns, and the utterance lengths are longer. For example, MultiWOZ 2.2 has lines more dispersed when the \emph{Reasoning Step} \textbf{$\phi(\text{step})$} is 2 compared to 1 as shown in the first row's first three columns in ~\autoref{tab:fine-grain}. We can also get a similar conclusion from \emph{Dialogue Turns} \textbf{$\phi(\text{turn})$} on dataset M2M-R+M. 10-14, 15-19, and 20+ are less dense than 0-9. As for \emph{Average Utterance Length} \textbf{$\psi(\text{len})$}, we can also find that for longer utterance lengths, CoTE-Coarse and CoTE-Refined improve more on three datasets. Quantitatively, our CoTE-Refined's average improvement margins compared to the strong baseline SDP on the reasoning steps 1, 2, and 3 are 1.55\%, 1.9\%, and 1.675\% for MultiWOZ 2.2, and 16.5\%, 23.4\% and 17.4\% for M2M-R+M, which exhibits better performance when steps are 2 and 3. On these observations, our approach is more effective in addressing intricate samples owing to the proposed method to simultaneously generate answers and CoT explanations.

\paragraph{O3: CoTE outperforms Summary.} We observe DS2, which generates slot values with an auxiliary summary of dialogue history, is also more robust in longer \textbf{$\phi(\text{step})$}, \textbf{$\phi(\text{turn})$} and \textbf{$\psi(\text{len})$} compared to SDP. For example, in M2M-R+M, DS2's performance is relatively stable in the line charts from left to right in each aspect. This observation shows that the summarization task used in DS2 also assists intricate slot value tracking. However, due to its overall lower performance than our CoTE variants, we can conclude that the summary is less effective than our multi-step chain-of-thought explanations.
\begin{table}[t]
    \centering \small
    \renewcommand\tabcolsep{3pt}
    \resizebox{0.95\linewidth}{!}{
    \begin{tabular}{l c c c}
        \toprule
         & MultiWOZ2.2 & M2M-R+M & WOZ 2.0 \\
        
        \midrule
        CoTE-Refined* & 56.6 & 87.0 & 87.2 \\
        CoTE-Refined & 57.5 & 91.6 & 90   \\
        \bottomrule
        \end{tabular}}
        \vspace{-7pt}
    \caption{Ablation study for CoTE explanations. * denotes that CoTE-Refined does not consider the explanations when training.}
    
    \label{tab:main-explain-ablation}
\end{table}
% \paragraph{O4: CoTE performs better in smaller datasets.} Our CoTE method shows even higher improvement margins in smaller datasets. For example, while SDP shows decent results in dataset MultiWOZ 2.2, it lags in smaller datasets such as M2M-R+M and WOZ2.0. As shown in~\autoref{tab:fine-grain}, the lines in these charts for M2M-R+M and WOZ2.0 are more sparsely distributed than MultiWOZ 2.2. 
% which could be attributed to the fact that the prompts used by SDP may not be easily comprehensible to pre-trained language models (PLMs). 

\subsection{Exp-IV: Is Explanation Effective?}

We conducted an ablation study to investigate how much gains the explanations bring.
We compared the performance before and after removing the explanations, and the results are shown in \autoref{tab:main-explain-ablation}.
The results show that the performance of CoTE drops a lot when the explanations are deleted. 
This suggests that the introduction of explanations in CoTE is effective.

\section{Conclusion}
\vspace{-3pt}
This work focuses on the dialog state tracking(DST) task. Our analysis of the data reveals that approximately 40\% of the samples within the DST task necessitate multi-step reasoning, and current models exhibit suboptimal performance when handling these instances. To mitigate this problem, we introduce CoTE-Coarse for DST, which concatenates relevant conversational turns associated with specific slots to serve as explanations when generating slot values. Moreover, to study whether fluent and high-quality reasoning content further enhances the model, we refine CoTE by GPT3 paraphrasing (CoTE-Refined). To explore where CoTE-Coarse/CoTE-refined works, we design a fine-grained evaluation framework for DST.
Experimental results in both full-dataset and low-resource settings demonstrate that our CoTE variants surpasses baselines in more intricate samples (e.g. requiring multi-step reasoning), and CoTE-Refined achieves additional performance enhancements.

% In this work, we conducted a comprehensive investigation of generative DST tasks and study how to effectively track the flow of dialogue states. We further improve schema-driven DST generation by introducing Chain-of-Thought Explanation to DST generation, and use GPT-3 to enhance the fluency and coherence of generated texts. Experiments and ablation studies have shown that the proposed model achieves state-of-the-art performance on multiple datasets, which demonstrates the effectiveness of our proposed methods. 

% For future work, further research could explore the effectiveness of Chain-of-Thought and in-context learning in low-shot scenarios, with a focus on the performance of supervised fine-tuning using extremely little data on pretrained models or large language models such as GPT-4 and LLaMA.

% \clearpage
\section*{Ethics Statement}
Dialogue state tracking has broad application prospects in dialog systems, emotional escort robots, intelligent assistants, etc. This work focuses on the way to construct explanations for generating accurate states. It plays a significant role in understanding multi-turn dialogues. All datasets we used in this work were privacy filtered and content moderated by the dataset authors~\cite{budzianowski-etal-2018-multiwoz, loshchilovdecoupled, mrksic-etal-2017-neural}. However, the way to use the large language models to generate explanations could produce harmful content and lead to inappropriate or biased understanding of dialogue histories.  Future work should take this into consideration.
% \section*{Limitations}
% The limitations of our works are listed below:
% (1) Due to limited computing resources, our experiments are restricted to a smaller model, namely T5-base. As a result, it remains to be seen how our idea will perform on larger models, such as LLaMA with 7B hyperparameters. We will explore it in the future.
% (2) Our model aims to tackle samples that involve multi-step reasoning or long dialogues. However, our method may lose the advantage when evaluating datasets primarily consisting of short dialogue samples and relying on single-step reasoning.
% (3) Our method needs to expand a sample to a predefined number of slots, which introduces complexities during the training process.

% \section*{References}\label{sec:reference}

\bibliography{anthology,custom}

\appendix

% \section{Example Appendix}
% \label{sec:appendix}

% This is an appendix.

\end{document}